\documentclass[runningheads]{llncs}
\usepackage[T1]{fontenc}
\usepackage{subfig}
\usepackage{graphicx}
\usepackage{xcolor}
\usepackage{hyperref}
\usepackage{soul}

\newcommand{\FZ}[1]{{\color{black} #1}}

\begin{document}
\title{Vision-Based Robotic Disassembly Combined with Real-Time MFA Data Acquisition}
\titlerunning{Vision-Based Robotic Disassembly Combined with MFA Data Acquisition}
%
\author{Federico Zocco\orcidID{0000-0002-6631-7081} \and
Maria Pozzi\orcidID{0000-0002-2390-1907} \and
Monica Malvezzi\orcidID{0000-0002-2158-5920}}
\authorrunning{F. Zocco et al.}
%
\institute{Department of Information Engineering and Mathematics, University of Siena, Italy, Via Roma, 56, 53100.\\
\email{federico.zocco.fz@gmail.com, maria.pozzi@unisi.it, monica.malvezzi@unisi.it}
}
\maketitle              
\begin{abstract}
Stable and reliable supplies of rare-Earth minerals and critical raw materials (CRMs) are essential for the development of the European Union. Since a large share of these materials enters the Union from outside, a valid option for CRMs supply resilience and security is to recover them from end-of-use products. Hence, in this paper we \FZ{present the preliminary phases of} the development of real-time visual detection of PC desktop components running on edge devices to simultaneously achieve two goals. The first goal is to perform robotic disassembly of PC desktops, where the adaptivity of learning-based vision can enable the processing of items with unpredictable geometry caused by accidental damages. We also discuss the robot end-effectors for different PC components with the object contact points derivable from neural detector bounding boxes.
The second goal is to provide in an autonomous, highly-granular, and timely fashion, the data needed to perform material flow analysis (MFA) since, to date, MFA often lacks of the data needed to accurately study material stocks and flows. The second goal is achievable thanks to the recently-proposed synchromaterials, which can generate both local and wide-area (e.g., national) material mass information in a real-time and synchronized fashion.   
\end{abstract}
\section{Introduction}
Modern societies provide products and services of increasing performance and customization. Digital technologies are the fuel of such a modernization. While the benefits of digitalization are clear and tangible, the way in which digital systems are currently designed and used is intrinsically unsustainable in the long term. This is because: (1) as of today, digital systems are made of non-renewable resources, e.g., critical raw materials (CRMs) \cite{CRMs}, which makes their manufacturing uncertain in the future; and (2) digital systems contribute to the generation of waste electrical and electronic equipment (WEEE), whose management is of increasing concern due to the large volumes. In numbers, the UN's Global E-waste Monitor recorded that 62 million tonnes of WEEE was produced in 2022, that is an 82\% increase compared to 2010 and is on track to reach 82 million tonnes in 2030, while just 1\% of rare-Earth element (REE) demand was met by WEEE recycling \cite{UN-GEM}. A further source of uncertainty for the future production of digital systems in the European Union is that, in 2024, the EU imported 46.3\% of REEs from China (first supplier) and 28.4\% from Russia (second supplier) \cite{EU-REEs}. A reduction of the import of REEs would enhance the resilience of the EU, especially considering the current geopolitical situation.   

The paradigm of a circular economy (CE) is gaining interest to mitigate the unsustainable factors mentioned above \cite{potting2017circular}. The urgency of a shift from linearity to circularity is visible in the ongoing tensions between nations to secure the access to non-renewable resources, e.g., CRMs and oil \cite{baranowski2025russian}. The fundamental practices of CE are, for example, the reduction of CRM use, the reuse and repair of products to extend their life-cycle, the replacement of CRMs with renewable alternatives, and the recovery of CRMs from end-of-use products as effectively as possible (a.k.a. urban mining) \cite{potting2017circular}. 

One of the main factors hindering the transition to a circular economy are the high operational costs of disassembly and repair operations. Robots have the potential to accelerate the transition to CE since they can reduce the costs for carrying out those tasks if implemented on a large scale \cite{zocco2024unification,zocco2025towards}.

As well as increasing the implementation of circular practices such as disassembly, a transition to CE requires also to accurately map and study the dynamics of material transfers across their entire life cycle in order to identify at which stages interventions are needed to reduce the waste streams and the extraction of non-renewable resources. These system-level studies are carried out using material flow analysis (MFA) \cite{myers2025material}. Yet, to date MFA lacks of data to perform accurate models and simulate what-if scenarios \cite{mason2025bayesian}. In this context, this paper makes the following \FZ{\textbf{contributions}: \textbf{(1)}} To address the lack of data in MFA, we illustrate the principles for real-time generation of MFA data via vision-based synchromaterials \cite{zocco2025synchronized} in robotic disassembly (see Section \ref{sub:robotMFAdata}); \FZ{\textbf{(2)}} To progress in robotic processing of WEEE, we begin the development of vision systems for PC desktop disassembly and discuss the robot end-effectors for different PC components, where the object contact points are derivable from neural detector bounding boxes (see Section \ref{sub:visionInDisassembly}).

\section{Related Work}
Several works on robotic disassembly have been proposed over the years. A literature review on robotic disassembly was provided by Hjorth \emph{et al.} \cite{hjorth2022human} with a focus on human-robot collaboration for non-destructive disassembly, by Poschmann \emph{et al.} \cite{poschmann2020disassembly} with a focus on the use of robots for the automation of disassembly operations, by Ameur \emph{et al.} \cite{ameur2025future} with a focus on AI techniques enabling disassembly such as computer vision, and by Lee \emph{et al.} \cite{lee2024review}, which points-out that ``[...] While fully automating disassembly is not economically viable given the intricate nature of the task, there is potential in using human–robot collaboration [...]''. In this paper, we begin the development of vision-based robotic disassembly of PC desktops (see Section \ref{sub:visionInDisassembly}).

Robotic disassembly is one of the many stages of the life cycle of a material or product; examples of other stages are material extraction, transportation, assembly of single, and then, multiple components, and product use. Thus, an effective allocation and distribution of natural resources requires to look at the whole material life-cycle to identify and mitigate the losses, i.e., the waste streams, and also the stages with higher dependence on CRMs \cite{zocco2024unification}. This is the reason of the importance of MFA in understanding and studying the supply-chains in order to quantify the material circularity \cite{baars2022quo}. MFA is currently used by the European Commission \cite{RMIS-MFA}, by the National Interdisciplinary Circular Economy Research (NICER) programme launched in the UK in 2021 \cite{myers2025material}, and by the Australia's National Science Agency \cite{CSIRO-MFA}. Since a major limitation of existing MFA studies is the lack of material stocks-and-flows data \cite{mason2025bayesian}, in this paper we introduce the use of robots as an MFA data source (see Section \ref{sub:robotMFAdata}).

\section{Robots for Disassembly and MFA Data}
\subsection{Real-Time at-the-Edge Vision in Robotic Disassembly}\label{sub:visionInDisassembly}
While neural networks have become the foundations of modern computer vision, due to their computational cost they are not always the best option in robotic operations. Robotic disassembly of end-of-use items is an operation that can significantly benefit from neural vision systems given the unpredictable geometry of waste items caused by physical damages. Hence, we began the assessment of neural networks for real-time object detection executed on an NVIDIA Jetson Nano to recognize the components of end-of-use PC desktops. CRMs are contained in cables and in motherboards \cite{kohl2018physical}, and hence, their reuse or recycling is essential to reduce the dependence on CRMs. 

\begin{figure}
\subfloat[\label{fig:Labelling1}]{
  \includegraphics[width=0.5\textwidth]{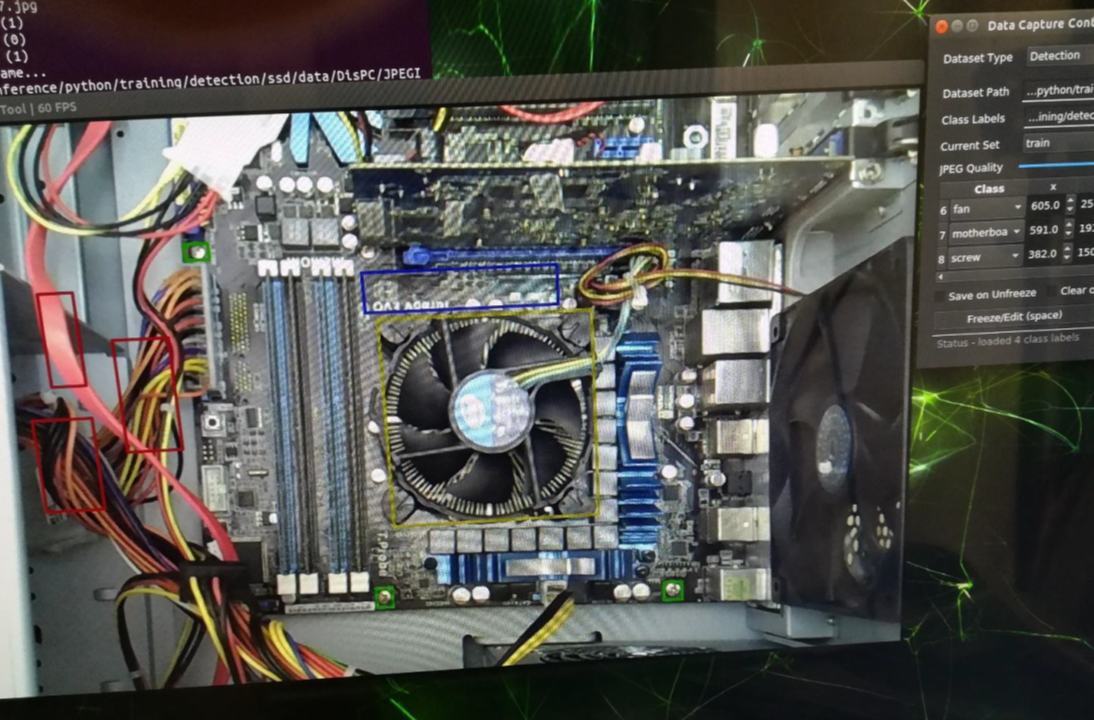}
  }
\subfloat[\label{fig:Labelling2}]{
  \includegraphics[width=0.43\textwidth]{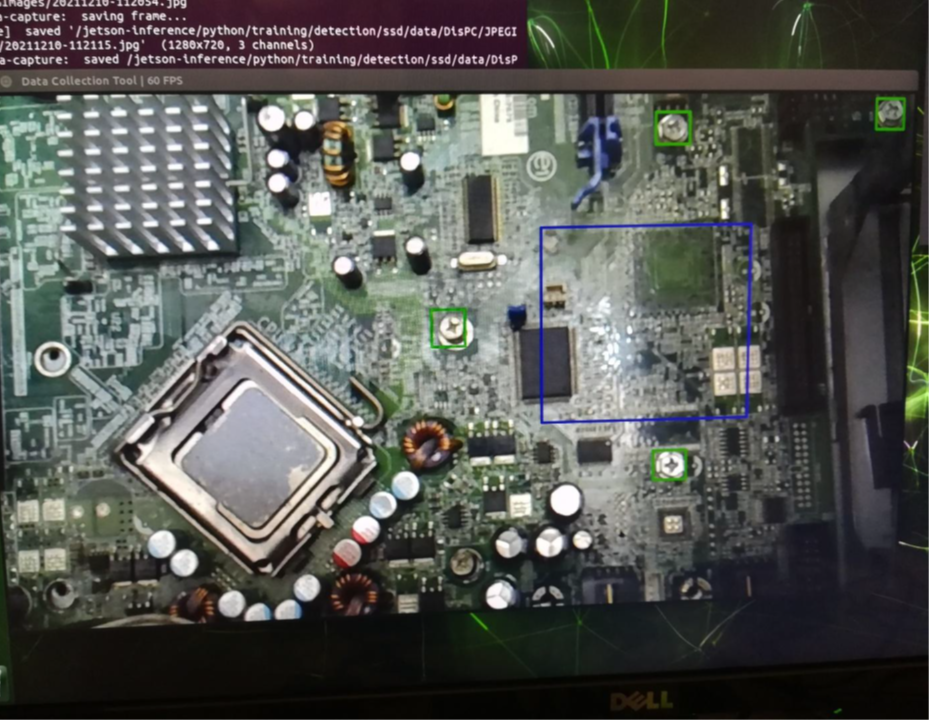}
  }
\caption{Annotation of PC desktop components considering different PC models with different end-of-use conditions. \FZ{Colors of bounding boxes: blue for the motherboard, yellow for the fan, red for the cables, and green for the screws.}}
\label{fig:Labelling}
\end{figure}

\textbf{Dataset preparation:} We collected and annotated 601 images using the NVIDIA annotation tool \cite{NvidiaODtutorial} and the Pascal VOC format. Realistic RGB sample images from three models of PC desktops with different end-of-use conditions were taken using a \FZ{Logitech C270} webcam. \FZ{The images were labeled using} 4 labels: cable, screw, fan, and motherboard. Specifically, we annotated: 
\begin{itemize}
\item{170 images with all the 4 labels taken from the first model of PC yielding a total of 8 bounding boxes (Fig.\ref{fig:Labelling1});} 
\item{80 images with 2-3 boxes for the screws, 1 box for the fan, and 1 box for the motherboard;} 
\item{80 images taken after the removal of the screws from the PC model used previously. Hence, no bounding boxes for screws were in those samples.}
\item{80 images taken from a second model of PC desktop considering the motherboard, groups of cables, and 2-4 screws.}
\item{40 images taken from a third model of PC considering 4 groups of cables.}
\item{151 images from the second model of PC with bounding boxes for the motherboard and the screws (Fig. \ref{fig:Labelling2}).}
\end{itemize}

\textbf{\FZ{Training and testing:}} We fine-tuned two neural detectors pretrained on the MS COCO dataset using all the 601 samples, namely, \emph{mobilenet-v1} and \emph{mobilenet-v2} \cite{NvidiaODtutorial}. Both models were fine-tuned for 120 epochs using the stochastic gradient descent optimizer and the following hyperparameters: base learning rate of 0.001, batch size of 2, momentum of 0.9, gamma of 0.1, weight decay of 0.0005, and without frozen layers. \FZ{All the 601 samples were used for training and validation.} 
The validation losses during training are reported in Table \ref{tab:validationLosses} \FZ{(quantitative model evaluation).} 

\begin{figure}
\subfloat[\label{fig:}]{
  \includegraphics[width=0.48\textwidth]{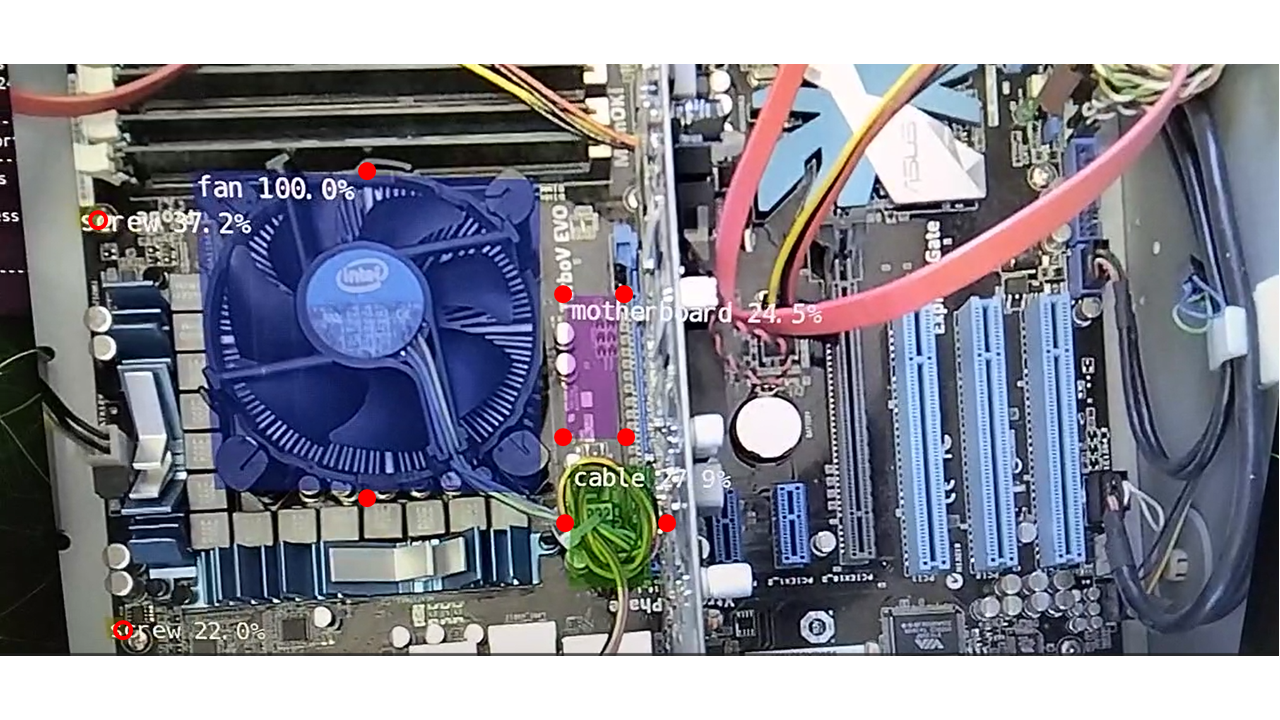}
  }
\subfloat[\label{fig:}]{
  \includegraphics[width=0.48\textwidth]{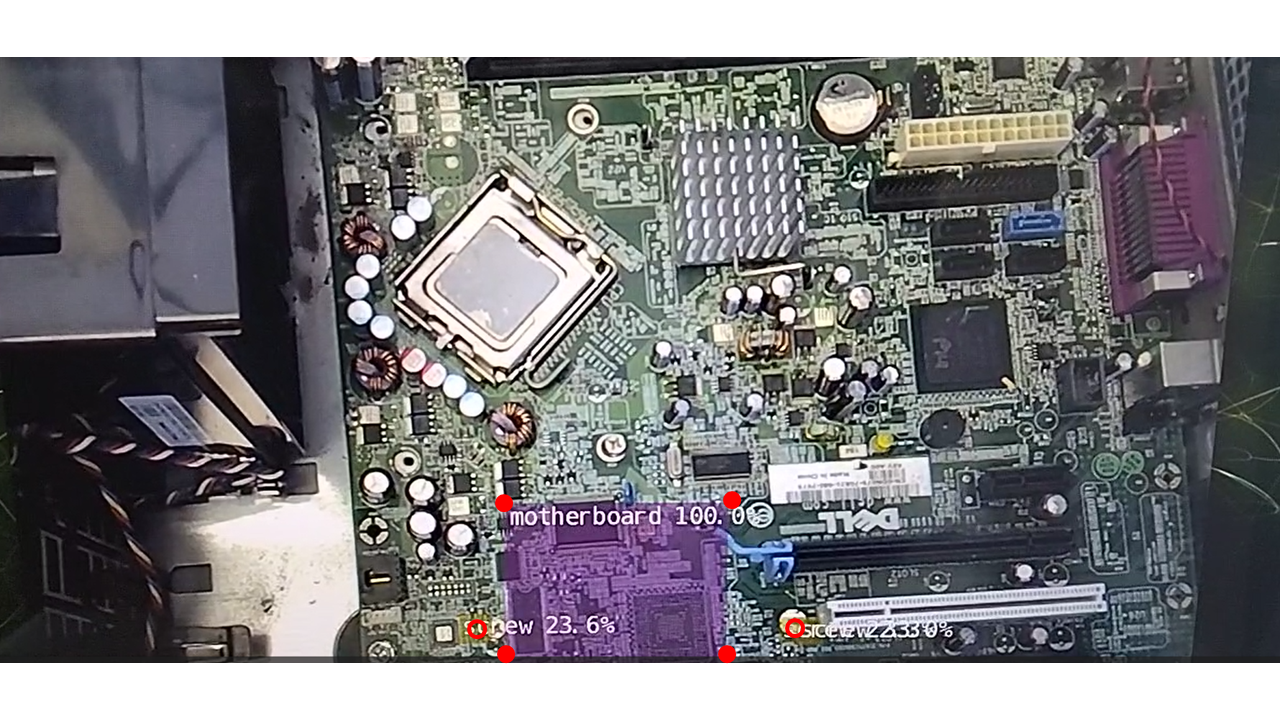}
  }
\caption{Bounding boxes generated in real-time by the detectors. The red dots and circles can be computed from the bounding boxes for positioning the grippers and tools needed to extract the PC components.}
\label{fig:InferenceGraspPoints}
\end{figure}

\FZ{Subsequently, a trained model was tested by running it to process, in real-time, the frames captured by the webcam. Since the PC case position was manually changed during the webcam capture, the frames seen by the model during this test were different from those in the training set. Examples of model outputs during this test are shown in Fig. \ref{fig:InferenceGraspPoints} (qualitative model evaluation)}. The elements in red are manually added to Fig. \ref{fig:InferenceGraspPoints} \FZ{to show how the bounding boxes generated by the neural detectors could potentially be used with different robot end-effectors.} Specifically: (1) the two dots at the edges of the fan and the cable boxes indicate the \FZ{antipodal} grasping points of a fingered gripper (e.g., \cite{pozzi2024soft}); (2) the four dots at the vertices of the motherboard bounding box indicate the vertices of the flat surface \FZ{which could be used} for \FZ{placing a suction gripper over the motherboard and grasp it} \cite{zhakypov2018origami}; and (3) the circles on the screws indicate the position of the magnetic-head screwdriver needed for unscrewing them.    
\begin{table}
\centering
\caption{Validation loss, regression loss, and classification loss for the detectors \emph{mobilenet-v1} (\emph{mobilenet-v2}).}
\label{tab:validationLosses}
\begin{tabular}{c@{\hskip 0.2in}c@{\hskip 0.2in}c@{\hskip 0.2in}c} 
Epoch & Validation loss & Val. regression loss & Val. classification loss\\ 
\hline
1 & 5.2\,(5.8) & 2.7\,(3.0) & 2.5\,(2.8) \\
40 & 2.0\,(2.1) & 0.6\,(0.7) & 1.3\,(1.3) \\
120 & 1.5\,(1.5) & 0.4\,(0.4) & 1.1\,(1.1) \\
\hline
\end{tabular}
\end{table}

\subsection{Robots for Wide-Area Real-Time Material Monitoring}\label{sub:robotMFAdata}
MFA studies are essential to enhance the allocation of natural resources, minimize the generation of waste, and increase the material efficiency of target areas, e.g., a city or a country \cite{mason2025bayesian}. However, MFA is a methodology that requires large volumes of data reporting the amount and the type of materials at different locations and times. In practice, MFA suffers from a lack of data \cite{mason2025bayesian}. 

\begin{figure}
    \vspace{-0.5cm}
    \centering
\includegraphics[width=\textwidth]{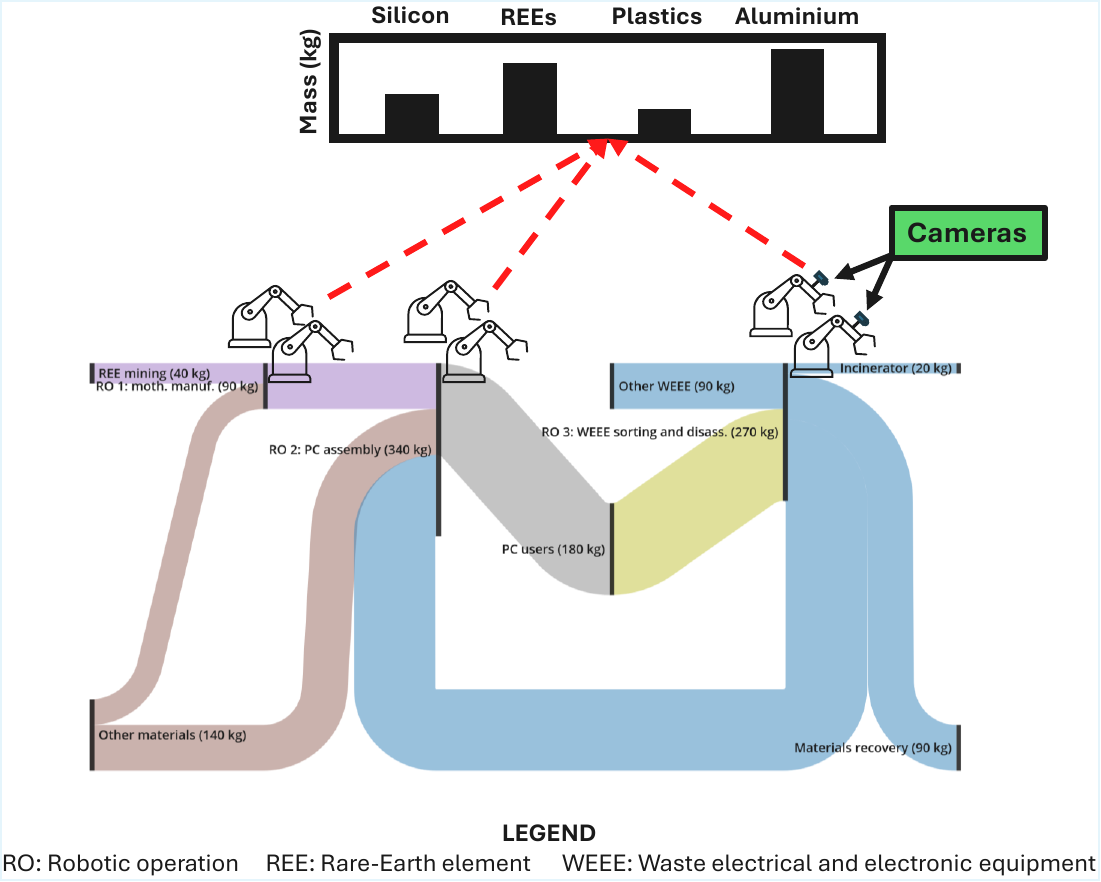}
    \caption{The data for an MFA Sankey diagram \cite{mason2025bayesian} can be generated via synchromaterials \cite{zocco2025synchronized} provided by the robots performing assembly, sorting, and disassembly. Computationally-intensive vision systems are justifiable in the case of robotic end-of-use processing (i.e., sorting and disassembly operations) since the geometries of waste items are unknown. \FZ{Amounts and materials are merely indicative.}
    }
    \label{fig:RobotsWithSynchr}
    \vspace{-0.5cm}
\end{figure}

To date, robots are not used to feed MFA studies with the missing data even though they could work  as depicted in Fig. \ref{fig:RobotsWithSynchr}, where a Sankey diagram capturing the material flows \cite{mason2025bayesian} is combined with robotic operations.   
Specifically, the Sankey diagram captures the transfers of material masses from minining to manufacturing, to use, to disassembly, and to incineration. In the case of disassembly covered in Section \ref{sub:visionInDisassembly} and indicated with ``RO 3'' in Fig. \ref{fig:RobotsWithSynchr}, the neural detector \emph{mobilenet} can inform the robot about the PC components to perform grasping tasks and, simultaneously, send material information in real-time for updating MFA Sankey digrams  \cite{zocco2025synchronized}. The mass and type of material can also be provided by assembly robotic operations, namely, by ``RO 1'' for the motherboard manufacturing and ``RO 2'' for the PC desktop assembly. In assembly operations, the constituent materials of the final product are known, and hence, the computationally-intensive vision systems covered in Section \ref{sub:visionInDisassembly} are unnecessary, e.g., when the $n$-th assembly step is ongoing, the robot sends the mass and type of materials of the product assembled until that moment; the information is updated during the step $n$+1, step $n$+2, and so on until the assembly is complete. The material information transmitted by the different robotic stages of the material life cycle, i.e., assembly, disassembly, and waste sorting, is added up at every sampling time, say, every minute, to inform about the material mass in a certain area of interest, e.g., a city or a country (yielding the bar chart at the top of Fig. \ref{fig:RobotsWithSynchr}). Since the material information must be synchronized in time to make consistent calculations, the name ``synchromaterials'' is used \cite{zocco2025synchronized}.

\section{Conclusion}

In this paper, we report on the initial assessment of vision-based robotic disassembly of PC desktops combined with material mapping for MFA studies. The neural detectors can detect the PC components, yet misdetections or missing detections occur mainly because of the limited size of the dataset collected for training (601 samples). These detection errors affect also the material mapping accuracy since they feed MFA with inaccurate material measures, i.e., inaccurate mass or material type. Hence, future work will consist of improving the accuracy of the neural detectors by increasing the dataset size and considering the use of vision transformers as they are showing very promising results in the last years~\cite{li2022exploring}.

\section*{Acknowledgment}
The authors gratefully thank Davide Rossi, Department of Information Engineering and Mathematics, University of Siena, Italy, for providing the PC desktops and for the valuable discussions.

%
%
%
\bibliographystyle{splncs04}
\bibliography{References}

\end{document}